\documentclass{article}

\usepackage[nonatbib, final]{neurips_2023_gaied}

\usepackage[utf8]{inputenc} 
\usepackage[T1]{fontenc}    
\usepackage{hyperref}       
\usepackage{url}            
\usepackage{booktabs}       
\usepackage{amsfonts}       
\usepackage{nicefrac}       
\usepackage{microtype}      
\usepackage{xcolor}         

\usepackage{algorithm}
\usepackage{times}
\usepackage{latexsym}
\usepackage{algorithm}
\usepackage{multirow}
\usepackage{algpseudocode}
\usepackage{amsmath}
\usepackage{makecell}
\usepackage{graphicx} 
\usepackage{enumitem}
\usepackage{subfigure}
\usepackage[framemethod=TikZ]{mdframed}
\mdfdefinestyle{MyFrame}{
    outerlinewidth=1pt,
    roundcorner=4pt,
    innertopmargin=\baselineskip,
    innerbottommargin=\baselineskip,
    leftmargin=5pt,
    innerrightmargin=15pt,
    innerleftmargin=20pt,
    backgroundcolor=gray!20!white,
}

\title{AuthentiGPT: Detecting Machine-Generated Text via Black-Box Language Models Denoising}

\author{%
  Zhen Guo \\
  MIT EECS\\
  Cambridge, MA 02139 \\
  \texttt{zguo0525@mit.edu} \\
  \And
  Shangdi Yu \\
  MIT EECS \\
  Cambridge, MA 02139 \\
  \texttt{shangdiy@mit.edu} \\
}

\begin{document}

\maketitle

\begin{abstract}
Large language models (LLMs) have opened up enormous opportunities while simultaneously posing ethical dilemmas. One of the major concerns is their ability to create text that closely mimics human writing, which can lead to potential misuse, such as academic misconduct, disinformation, and fraud. To address this problem, we present AuthentiGPT, an efficient classifier that distinguishes between machine-generated and human-written texts. Under the assumption that human-written text resides outside the distribution of machine-generated text, AuthentiGPT leverages a black-box LLM to denoise input text with artificially added noise, and then semantically compares the denoised text with the original to determine if the content is machine-generated. With only one trainable parameter, AuthentiGPT eliminates the need for a large training dataset, watermarking the LLM's output, or computing the log-likelihood. Importantly, the detection capability of AuthentiGPT can be easily adapted to any generative language model. With a 0.918 AUROC score on a domain-specific dataset, AuthentiGPT demonstrates its effectiveness over other commercial algorithms, highlighting its potential for detecting machine-generated text in academic settings.
\end{abstract}

\section{Introduction}
Large Language Models (LLMs) have significantly transformed the field of artificial intelligence, offering vast opportunities, but also raising important ethical questions. One of the main issues is their capacity to generate text that closely resembles human writing. Given their extensive scale and accessibility, if misused, LLMs could not only amplify harms such as disinformation and fake news~\cite{NEURIPS2019_3e9f0fc9}, but also undermine academic integrity by assisting unauthorized content generation like essay-writing for students. Such instances highlight the need to differentiate between content generated by machines and that created by humans~\cite{gehrmann2019gltr, ippolito-etal-2020-automatic, crothers2022machine, dwivedi2023so, liebrenz2023generating, he2023mgtbench}.

In the past, various methods have been developed to tackle the challenge of detecting machine-generated text~\cite{badaskar-etal-2008-identifying, beresneva2016computer, shah2020comparative, jawahar2020automatic, mitrovic2023chatgpt, sadasivan2023can, chakraborty2023possibilities, tang2023science}. One common strategy involves training separate classifiers on datasets containing both human and machine-generated texts with labels~\cite{sebastiani2002machine, islam2023distinguishing}. However, the effectiveness of these supervised classifiers often requires a large amount of training data, therefore increasing the cost associated with the training procedure.
Another set of evaluation techniques relies on computing the log-likelihood of the language models~\cite{mitchell2023detectgpt, deng2023efficient}. 
However, obtaining the parameters of these models can often be challenging, as they may be inaccessible through chat interfaces or commercial APIs.
Watermarking LLM outputs could also offer proactive detection of machine-generated text, maintaining trust and transparency~\cite{kirchenbauer2023watermark,kirchenbauer2023reliability,munyer2023deeptextmark}. However, this technique needs a balance between imperceptibility and detectability, while also addressing potential attacks on the watermarks to ensure the integrity of the detection process~\cite{krishna2023paraphrasing}.

\begin{table}
  \caption{AUROC scores on the aggregated PubMedQA and GPT generated QA datasets. AuthentiGPT outperforms zero-shot GPT-3.5 and GPT-4, and surpasses GPTZero and Originality.AI.}
  \label{tab:auroc-scores}
  \centering
  \begin{tabular}{lc}
    \toprule
    \textbf{Methods} & \textbf{AUROC Scores} \\
    \midrule
    GPT-3.5 (zero-shot) & 0.721 \\
    GPT-4 (zero-shot) & 0.577 \\
    GPTZero & 0.797 \\
    Originality.AI & 0.906 \\
    AuthentiGPT & \textbf{0.918} \\
    \bottomrule
  \end{tabular}
\end{table}

In this paper, we introduce AuthentiGPT, a novel classification algorithm designed to distinguish between machine-generated and human-written text. Unlike previous methods that rely on extensive labeled data, watermarking, or log-likelihood calculations, AuthentiGPT leverages the language model itself for detection. With the assumption that human-written text resides outside the distribution of machine-generated text~\cite{Holtzman2020The}, the algorithm first introduces synthetic noise to the input text and utilizes a black-box LLM to denoise the text at a noise level. Then, by comparing the denoised text with the original text at the semantic level, a lightweight classifier (with only one free parameter) that is trained on a small set of examples can effectively determine the text was generated by a machine or a human. Our experimental results (Table~\ref{tab:auroc-scores}) show that AuthentiGPT outperforms existing methods in detecting machine-generated content within PubMedQA and generated QA datasets, demonstrating the effectiveness of the algorithm. An important advantage of AuthentiGPT is its adaptability. As LLMs continue to improve, AuthentiGPT can be easily adapted to them with minimal effort and modification.

While expanding the capabilities of LLMs, it becomes increasingly crucial to address the ethical considerations surrounding their use. AuthentiGPT represents a valuable advancement in this regard and improves language models' responsible and ethical application.

\section{Related Works}

\subsection{Language Model-based Classifiers}
Classification tasks can utilize language models by incorporating them with classification layers to perform detections. A conventional classifier that is based on a language model usually operates by passing the language model's hidden states into a fully-connected layer, which subsequently handles the classification:
\begin{equation}
y = \text{softmax}(W \cdot \text{LM}(S) + b).
\end{equation}
In this equation, \(\text{LM}(S)\) is the hidden states of the language model for input senetence \(S\), which has been previously processed by a tokenizer. \(W\) and \(b\) are the parameters of the classifier, and \(y\) is the predicted label. The benefits of using a language model for classification allow the algorithms to capture complex language patterns and context, making them highly effective in understanding and classifying text data. However, depending on the dimension of \(W\), these classifiers often require a lot of training data and may struggle with the evolving complexity of larger models~\cite{liang2023gpt}.

\subsection{Log-likelihood (perplexity) Computation}

Perplexity is a measurement in information theory and is commonly used in language modeling to assess how well a probability model predicts a sample~\cite{moore2010intelligent}. For LLMs, perplexity measures the model's uncertainty in predicting the next word in a sentence. A lower perplexity indicates that the model is better at predicting the sequence of words, and thus, indicting a higher likelihood of machine-generated content~\cite{vasilatos2023howkgpt}. To compute perplexity, one needs to compute the log-likelihood of text being generated by a model. For input tokens $X = (x_0, x_1, \dots, x_t)$, we have

\begin{equation}
\text{PPL}(X) = \exp \left\{ {-\frac{1}{t}\sum_i^t \log p_\theta (x_i|x_{<i}) } \right\},
\end{equation}
where $\log p_\theta (x_i|x_{<i})$ is the log-likelihood of the i-th token conditioned on the preceding tokens $x_{<i}$ given the language model with parameters of $\theta$. More sophisticated method, such as DetectGPT, uses the log probability computed by the model with a curvature-based criterion to assess the probability of content being machine-generated~\cite{mitchell2023detectgpt}. Despite the prospective advantages of this method, it comes with computational demands and requires access to the language model's parameters, which may not be available in all scenarios. Additionally, its accuracy tends to diminish when faced with complex language models, limiting its overall effectiveness.

\subsection{Watermark Detection}

Watermark detection in machine-generated text involves pinpointing distinctive traits or patterns that suggest its origin or the model that produced it~\cite{kirchenbauer2023watermark}. However, watermark detection encounters obstacles such as the difficulty in recognizing consistent patterns due to the variability of text, safeguarding against adversarial attacks, ensuring watermark robustness without compromising the quality of text, and more. Overall, this field continues to evolve with the advent of more sophisticated watermarks, requiring further development for practical application.

\subsection{Black-box Detection with N-Gram Analysis}
A concurrent study explores the use of a black-box LLM through divergent N-gram analysis (DNA-GPT) for the detection of machine-generated text~\cite{yang2023dna}. They defined the DNA-GPT BScore by:
\begin{align}
    \text{BScore}(S, \Omega) &= \frac{1}{K}\sum^{K}_{k=1}\sum^{N}_{n=n_0}f(n) \nonumber \\
    &\times \frac{|\textit{n-grams}(S_k^{'}) \cap \textit{n-grams}(S_2)|}{|S_k^{'}||\textit{n-grams}(S_2)|},
\end{align}
where $S$ is the input text sentence, $\Omega$ is a set of sentences sampled from the LLM. $K$ is the sampling repetition, $f(n)$ is a weight function for different $\textit{n-grams}$, $S_2$ is a substring of $S$, and $S'_k$ are machine-generated sentences based on $S \setminus S_2$. The final classification threshold is determined by balancing the true positive rate and false positive rate. Though the underlining principle of DNA-GPT is similar to AuthentiGPT, AuthentiGPT used a embedding model, instead of n-gram, to extract the features in the input sentences, and utilized a combination of non-linear transformation and unsupervised clustering to determine the threshold, instead of heuristically picking one. The paper claims that the detection strategy is ``training-free", but balancing the true positive rate and false positive rate to determine the classification threshold still requires training examples.

\section{AuthentiGPT}

AuthentiGPT is an efficient detection algorithm that eliminates the requirement for a substantial training dataset, the application of watermarks on the LLM's output, or the computation of the log-likelihood. It does not need access to the language model's parameters and can accommodate changes in the LLM with minimal effort. Our algorithm is inspired by~\cite{graham2022denoising}, which addresses the problem of
out-of-distribution detection for images using denoising diffusion probabilistic models (DDPMs). By denoising an input image that has been noised to a range of noise
levels, the multi-dimensional reconstruction errors can be obtained and then used to classify out-of-distribution inputs. We apply a similar process to sentences with artificially added noise using an instruction-tuned language model.

On a high-level, AuthentiGPT operates under the assumption that \textit{human-written text resides outside the distribution of machine-generated text}. The algorithm first utilizes a black-box language model to denoise input text with artificially added noise. Then, a semantic comparison is performed between the denoised text and the original text to determine whether it lies within or outside the distribution. In the following, we outline the step-by-step process of our algorithm in Algorithm~\ref{alg:authenti}. Inputs to the algorithm are a black-box LLM, some sentences $S_{\text{test}}$ that we want to classify, some training sentences $S_{\text{train}}$, and the labels of these training sentences. It has two parameters, $\alpha$ (masking ratio) and $\beta$ (the number of repetitions). We will show that AuthentiGPT can be effective with 10 training samples.

\begin{algorithm}[h]
\caption{Detecting Machine-Generated Text}\label{alg:authenti}
\begin{algorithmic}[1]
\Procedure{GetSimilarity}{$S$, \texttt{LLM}}
\For{i in $[1, 2, \ldots, \beta]$}
    \State $M \gets \text{maskSentences}(S, \alpha)$
    \State $D \gets \text{denoiseSentences}(M, \texttt{LLM})$
    \State $E_S \gets \text{computeEmbeddings}(S)$
    \State $E_D \gets \text{computeEmbeddings}(D)$
    \State $D_{\text{sim}, i} \gets \text{getSimilarity}(E_S, E_D)$
\EndFor
\State \Return $\text{mean}([D_{\text{sim}, 1}, \dots,  D_{\text{sim}, \beta}])$
\EndProcedure
\Procedure{AuthentiGPT}{\texttt{LLM}, $S_{\text{test}}$, $S_{\text{train}}$, $\text{labels}$ }
\State $\mathcal{D}_{\text{train}} \gets $ GetSimilarity($S$, \texttt{LLM})
\State $\texttt{gm} \gets $ FindThreshold($\mathcal{D}_{\text{train}}$, labels)
\State $\mathcal{D}_{\text{sim}} \gets $ GetSimilarity($S_{\text{test}}$, \texttt{LLM})
\State \Return \texttt{gm}.classify($\mathcal{D}_{\text{sim}}$)
\EndProcedure
\end{algorithmic}
\end{algorithm}

\begin{algorithm}[h]
\caption{Determine classification threshold} \label{alg:classification}
\begin{algorithmic}[1]
\Procedure{FindThreshold}{$\mathcal{D}_{\text{sim}}$, labels}
\For{$\lambda$ in $[\lambda_1, \lambda_2, \ldots, \lambda_n]$}
\State $\tilde{\mathcal{D_\lambda}} \gets \text{Box-Cox}(\mathcal{D_{\text{sim}}}, \lambda)$
\State \texttt{gm$_\lambda$} = GaussianMixture($\tilde{\mathcal{D_\lambda}}$, n\_class=2)
\State \texttt{score$_\lambda$} = AUROC(\texttt{gm$_{\lambda}$}, labels)
\EndFor
\State \Return the \texttt{gm$_{\lambda}$} and corresponding $\lambda$ that yield the maximum AUROC score
\EndProcedure
\end{algorithmic}
\end{algorithm}
Our algorithm performs the following operations:
   \begin{itemize}[itemsep=1pt,parsep=1pt]
   \item Randomly masks a portion of the sentences $S$ determined by a ratio $\alpha$ to create $M$, a version of $S$ with added noise. 
   \item Denoises sentences $M$ with the language model with completion or instruction, yielding a denoised version, $D$, of the sentences. 
   \item To semantically compare the original and denoised sentences, the algorithm computes embeddings for both $S$ and $D$, denoted as $E_S$ and $E_D$, respectively, and then computes the cosine similarity, $D_{\text{sim}}$, between the embeddings of the original and denoised sentences.
   \item The process repeats $\beta$ times to allow statistical significance.
   \item The averaged similarity score is sent to the classifier \texttt{gm} for classification. We will explain later in this section how we obtain this classification model.
   \end{itemize}

In our implementation, the mask function is on the word-level with \texttt{<unk>} to represent each masked word, the black-box language model \texttt{LLM} is \texttt{gpt-3.5-turbo} and embeddings are computed by \texttt{text-embedding-ada-002} from OpenAI\footnote{\url{https://platform.openai.com/docs}}. Note that the testing sentence $S_{\text{test}}$ does not necessarily have to be generated by the input \texttt{LLM}; it can be generated by other language models. Averaged cosine similarity serves as a measure of how much the denoised sentences deviate from the original ones after the masking and denoising process. Higher similarity scores indicate that the denoised sentences closely resemble the original ones, thus a higher likelihood of machine-generated text. Conversely, lower similarity scores suggest greater divergence, thus a lower likelihood of machine-generated text. Importantly, computing cosine similarity does not require the parameters of the language model. This is critical because most advanced models are API-based and function as black boxes to users and developers.

To classify the two groups of similarity scores, we use Algorithm~\ref{alg:classification} to determine the classification boundary. The inputs are $\mathcal{D}_{\text{sim}}$, a list of training examples that contain similarities between human-written and machine-generated texts, and their corresponding labels. We use a combination of Box-Cox power transformation~\cite{sakia1992box} and Gaussian Mixture Model (GMM)~\cite{rasmussen1999infinite, reynolds2009gaussian} to perform a soft classification. The Box-Cox transformation helps normalize the data, while the GMM enables the identification of the classification boundary in a probabilistic manner. We prefer GMM over other clustering algorithms for its Gaussian distribution assumption, which aligns with our dataset's characteristics. The output of Algorithm~\ref{alg:classification} is the Box-Cox parameter $\lambda$ and its corresponding GMM that yields the maximum AUROC (Area Under the Receiver Operating Characteristic Curve) score~\cite{delong1988comparing} on the training set, which will be used to classify the test datasets. Since $\lambda$ is the only trainable parameter, Algorithm~\ref{alg:classification} is data-efficient. While our current approach requires minimal training data, transitioning to more advanced classification techniques might be beneficial if more training data is available.

\begin{figure*}[h!]
    \centering
    \includegraphics[width = \textwidth]{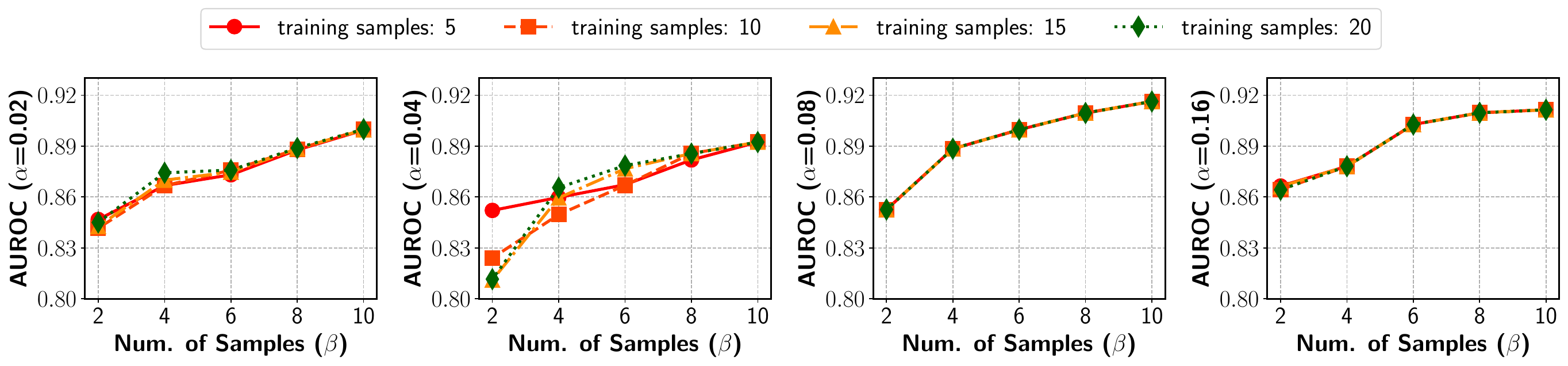}
    \caption{The AUROC scores of AuthentiGPT using different training samples and masking ratios. The x-axis has the number of averaging samples $\beta$. Each plot shows the scores using different masking ratios $\alpha$.}
    \label{fig:line_plot}
\end{figure*}

In our experiments, we evaluated our method ranging from 5 to 20 training examples. Further in the Section~\ref{sec:result}, we demonstrate that the number of training instances has a minimal impact on the performance of AuthentiGPT. Even with a mere 10 examples from both the original and GPT-generated PubMedQA datasets, the algorithm classifies effectively. This is unsurprising since the only trainable parameter for the algorithm is $\lambda$ and GMM is an unsupervised clustering model that requires no training. This highlights AuthentiGPT's remarkable proficiency in differentiating between human-written and machine-generated text. For Algorithm~\ref{alg:classification}, we utilize 100 $\lambda$ values for a grid search. The runtime for this subroutine is negligible when compared to the denoising process.

\section{Experiment Settings}

\subsection{Dataset}

Our dataset includes the original PubMedQA~\cite{jin2019pubmedqa} and machine-generated texts from PubMedQA. For evaluation, we use 80 out of 100 instances from each dataset. The rest of the 20 instances from each dataset are combined and used to determine the soft classification threshold. 

\textbf{Human-written texts} contain 100 original question and answer pairs sourced from PubMedQA. These QA pairs were created by humans and are considered to be reliable and accurate within the biomedical domain.

\textbf{Machine-generated texts} include 400 QA pairs that were generated by language models, specifically GPT-3.5 and GPT-4. Two sets of instructions were provided: one to rewrite existing QA pairs and another to generate new QA pairs using the 100 human-written QA pairs as references. These datasets are obtained from~\cite{guo2023dr}. Selected examples are shown in Section~\ref{sec:example}.

\subsection{Evaluation Metrics}

\textbf{Accuracy:} It measures the proportion of correct predictions made by a method for a data set that contains only one class (human-generated or machine-generated), considering both true positives and true negatives.

\textbf{AUROC:} The ROC curve is created by plotting the true positive rate against the false positive rate at various threshold settings. To compute the AUROC scores for different methods, we aggregate the classification results across all datasets. This approach allows us to obtain a comprehensive assessment of the algorithms' performance across the entire set of datasets.

\subsection{Baseline Methods}
\textbf{GPT-3.5 (Zero-shot):} This method, in a zero-shot setting, generates a binary response ("yes" or "no") indicating whether a given text is machine-generated or not using GPT-3.5. 

\textbf{GPT-4 (Zero-shot):} Similar to the zero-shot GPT-3.5 method, utilizing GPT-4 model instead.

\textbf{GPTZero\footnote{\url{https://gptzero.me/}}:} A commercially available classifier, trained to distinguish between human-written and machine-generated texts. It claims to be the most accurate AI detector across use cases.

\textbf{Originality.AI\footnote{\url{https://originality.ai/}}:} Another commercial classifier claimed to be the most accurate AI detection tool.

\section{Results}\label{sec:result}

\begin{figure*}[h!]
\centering
\begin{minipage}{0.495\textwidth}
   \includegraphics[width=1\linewidth]{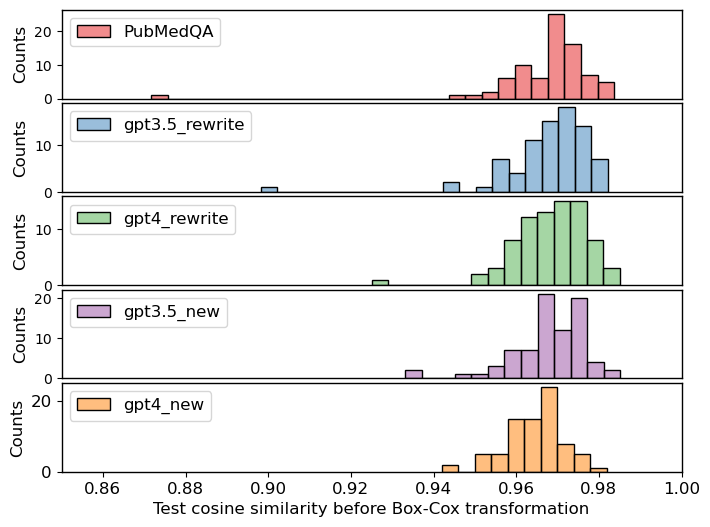}
\end{minipage}
\begin{minipage}{0.495\textwidth}
   \includegraphics[width=1\linewidth]{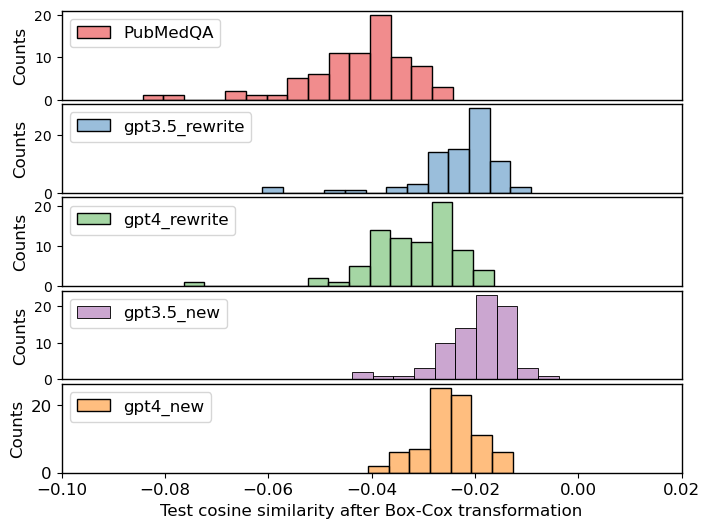}
\end{minipage}
\caption[]{The figure on the left is the histogram plots for the cosine similarity of the test datasets before the Box-Cox transformation. The figure on the right is the histogram plots for the cosine similarity of the test datasets after the Box-Cox transformation.}
\label{fig:dist1}
\end{figure*}

\begin{figure}[h!]
    \centering
    \includegraphics[width=0.65\columnwidth]{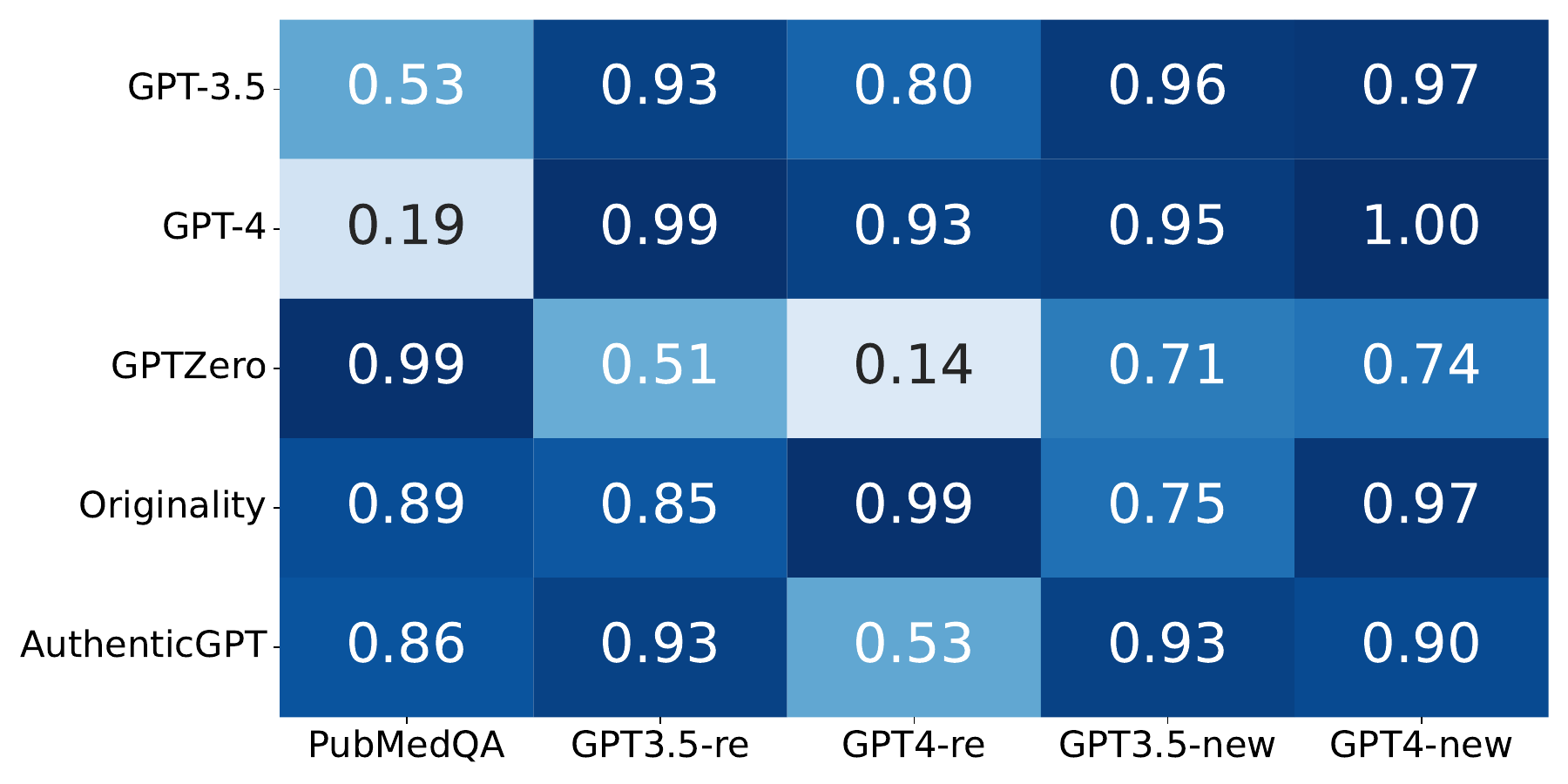}
    \caption{Accuracy on individual datasets. `GPT3.5-re` and `GPT4-re` are the datasets obtained by re-writing using GPT-3.5 and GPT-4 respectively. `GPT3.5-new` and `GPT4-new` are the datasets obtained by generating new QA pairs using GPT-3.5 and GPT-4 respectively. }
    \label{fig:performance}
\end{figure}

In Figure~\ref{fig:line_plot}, we present the AUROC scores of AuthentiGPT on the combined PubMedQA and GPT-generated QA dataset with different masking ratios and the number of training samples. As the masking ratio $\alpha$ increases, the performance of the algorithm also improves, reaching a plateau at around a masking ratio of 0.08. Intuitively, finding an optimal masking ratio requires meticulous tuning. A high masking ratio may result in a loss of crucial semantic information and thus limits the context for the LLM to work with, making the denoising task challenging. Consequently, the denoised sentences, whether machine-generated or human-written, would deviate significantly in semantics from their original versions regardless, making the classification task difficult. On the other hand, if the masking ratio is very small, the denoising task would be trivial. As a result, denoised sentences may remain semantically close to the original ones, which are also difficult to classify. Therefore, these considerations necessitate careful balance in the masking ratio for optimal performance for the algorithm. 

On the other hand, the influence of the number of training samples appears minimal except for $\alpha=0.04$. Again, this is unsurprising since the only trainable parameter for the algorithm is $\lambda$ and GMM is an unsupervised clustering model that requires no training.  A higher $\beta$ consistently improves the performance. In other words, more samples increase the statistical significance of the classification results. However, it's crucial to note that the runtime for the algorithm is linearly correlated with $\beta$. In the rest of the experiments, we will use a masking ratio ($\alpha$) of 0.08, number of averaging sample ($\beta$) of 10, and 20 training samples for AuthentiGPT. 

In Figure~\ref{fig:dist1}, we show the histogram of the cosine similarity between the original text and the denoised text from the black-box language model \texttt{gpt-3.5-turbo} with $\alpha$ of 0.08 and $\beta$ of 10. The hyperparameter $\lambda$ is of 0.5, which will also be used in Figure~\ref{fig:performance}. After applying the Box-Cox transformation, the dataset distribution resembles a Gaussian distribution, and the separation between different dataset distributions becomes more discernible. This statistical technique is crucial for our final classification using GMM clustering.

Table~\ref{tab:auroc-scores} shows the AUROC scores for various detection methods on the combined PubMedQA and GPT-generated QA dataset. Notably, AuthentiGPT demonstrates superior performance compared to zero-shot GPT-3.5 and GPT-4, and slightly surpasses Originality.AI, a high-end commercial classifier. GPTZero, another commercial classifier, exhibits intermediate performance.

In Figure~\ref{fig:performance}, we present a comparative analysis of the methods on each individual dataset. The performance is evaluated using the accuracy metric. 
We observe that some methods exhibit a tendency to classify sentences into one class. Specifically, GPT-3.5 and GPT-4 tend to classify sentences as machine-generated while GPTZero tends to classify sentences as human-written. As a result, though GPT-3.5 and GPT-4 correctly classify the machine-generated sentences, they fail at correctly classifying human-written sentences (original PubMedQA). Similarly, though GPTZero correctly classifies human-written sentences, it has much lower accuracy in classifying machine-generated sentences.
Across tasks, Originality.AI is consistently good except for \texttt{gpt3.5-new} task. AuthentiGPT performs well across the board, with the exception in \texttt{gpt4\_rewrite}. These variations can be attributed to the inherent complexities and subtle distinctions within each dataset. For GPTZero and AuthentiGPT, classifying GPT rewrites is more challenging than new QA tasks, potentially due to the retention of the human-written text style and semantics in the rewrite datasets~\cite{mitrovic2023chatgpt}. While tasks involving generated new QAs are comparatively easier as the machine-generated patterns become more distinguishable.

\section{Limitations and Future Work}

Although AuthentiGPT outperforms all other methods on the aggregated PubMedQA datasets, it exhibits shortcomings in individual datasets, particularly in \texttt{gpt4\_rewrite}. In addition, several limitations must be acknowledged. Firstly, the extent to which the algorithm's effectiveness applies beyond the biomedical field is unclear. Future work should focus on assessing its generalizability across a variety of domains. Secondly, our assumption that human-written text lies outside the distribution of machine-generated text may not always hold true. This could potentially impact the algorithm's performance. Finally, the risk of false positives in detection, which could undermine trust in assessment tools, is a critical but unaddressed concern. These considerations highlight the need for further optimization of the method, such as including additional black-box language models for ensemble averaging or using neural classifiers with more training samples.

\section{Conclusion}

In conclusion, AuthentiGPT offers a novel approach for detecting machine-generated text using small amount of training data. As language models continue to evolve and advance, the importance of tools like AuthentiGPT would become increasingly apparent. Those tools may play a crucial role in ensuring the ethical and responsible application of these models.

\bibliographystyle{unsrt}  
\bibliography{ref}  
\appendix
\section{Examples from our datasets}\label{sec:example}

In this section, we show one example from each of our five datasets. The \texttt{PubMedQA} is included in prompting language models to rewrite or generate a new question-answering pair in the biomedical domain.\\

\begin{mdframed}[style=MyFrame]
    \texttt{PubMedQA}\\
    \textbf{Question}: Syncope during bathing in infants, a pediatric form of water-induced urticaria? context: Apparent life-threatening events in infants are a difficult and frequent problem in pediatric practice. The prognosis is uncertain because of risk of sudden infant death syndrome.Eight infants aged 2 to 15 months were admitted during a period of 6 years; they suffered from similar maladies in the bath: on immersion, they became pale, hypotonic, still and unreactive; recovery took a few seconds after withdrawal from the bath and stimulation. Two diagnoses were initially considered: seizure or gastroesophageal reflux but this was doubtful. The hypothesis of an equivalent of aquagenic urticaria was then considered; as for patients with this disease, each infant's family contained members suffering from dermographism, maladies or eruption after exposure to water or sun. All six infants had dermographism. We found an increase in blood histamine levels after a trial bath in the two infants tested. The evolution of these "aquagenic maladies" was favourable after a few weeks without baths. After a 2-7 year follow-up, three out of seven infants continue to suffer from troubles associated with sun or water. answer: "Aquagenic maladies" could be a pediatric form of the aquagenic urticaria. \\\textbf{Answer}: the answer to the question given the context is yes
\end{mdframed}

\begin{mdframed}[style=MyFrame]
    \texttt{gpt3.5-rewrite}\\
    \textbf{Question}: Can the use of a new medication reduce symptoms of depression in adults with major depressive disorder? Context: Major depressive disorder (MDD) is a common mental health condition that can significantly affect a person's quality of life. Current treatments for MDD include therapy and medication, but many patients do not respond well to these options or experience side effects. A new medication, called WXY-19, has been developed that targets a specific neurochemical pathway involved in depression. A clinical trial was conducted with 100 adults diagnosed with MDD. Participants were randomly assigned to receive either WXY-19 or a placebo for 12 weeks. Results showed that participants who received WXY-19 had a significant reduction in depressive symptoms compared to those who received the placebo. \\
    \textbf{Answer}: Yes, the use of WXY-19 can reduce symptoms of depression in adults with major depressive disorder.
\end{mdframed}

\begin{mdframed}[style=MyFrame]
    \texttt{gpt4-rewrite}\\
    \textbf{Question}: Is syncope during bathing in infants a potential pediatric form of water-induced urticaria? context: Apparent life-threatening events in infants pose a challenging and common issue in pediatric practice, with an uncertain prognosis due to sudden infant death syndrome risks. Over six years, eight infants aged 2 to 15 months experienced similar symptoms while bathing: they became pale, hypotonic, motionless, and unresponsive upon immersion, recovering within seconds after being removed from the bath and stimulated. Initial diagnoses considered were seizures or gastroesophageal reflux, but these were doubtful. The possibility of a pediatric equivalent to aquagenic urticaria was then explored; like patients with this condition, each infant's family had members with dermographism, ailments, or rashes triggered by water or sun exposure. All six infants demonstrated dermographism. In two tested infants, blood histamine levels increased after a trial bath. The progression of these "aquagenic ailments" improved after a few weeks without bathing. Following a 2-7 year follow-up, three of the seven infants continued to experience issues related to sun or water exposure. \\
    \textbf{Answer}: Based on the context, it appears that "aquagenic ailments" might be a pediatric form of water-induced urticaria, so the answer to the question is yes.
\end{mdframed}

\begin{mdframed}[style=MyFrame]
    \texttt{gpt3.5-new}\\
    \textbf{New question}: Is there a correlation between maternal smoking during pregnancy and the risk of sudden infant death syndrome (SIDS)?Context: Sudden infant death syndrome (SIDS) is the unexpected and sudden death of an apparently healthy infant, typically during sleep. It is the leading cause of death in infants between 1 month and 1 year of age. Several factors have been identified as risk factors for SIDS, including maternal smoking during pregnancy. Maternal smoking is known to have negative effects on fetal development, and it has been suggested that it may also increase the risk of SIDS. \\
    \textbf{Answer}: Yes, maternal smoking during pregnancy has been identified as a risk factor for sudden infant death syndrome (SIDS).
\end{mdframed}

\begin{mdframed}[style=MyFrame]
    \texttt{gpt4-new}\\
    \textbf{New question}: Can genetic predisposition to dermographism play a role in infants experiencing syncope during bathing?context: A study investigated eight infants aged 2 to 15 months who were admitted over a period of 6 years with similar symptoms during bathing. These infants became pale, hypotonic, still, and unreactive upon immersion in water, and recovered within seconds once removed from the bath and stimulated. Initial diagnoses considered were seizure or gastroesophageal reflux, but these were deemed unlikely. Researchers then considered the possibility of a pediatric form of aquagenic urticaria. Each infant's family had members with dermographism, skin reactions to water or sun exposure, and all six infants tested had dermographism. An increase in blood histamine levels after a trial bath was observed in two infants tested. After a few weeks without baths, the condition improved, but after a follow-up of 2-7 years, three out of seven infants continued to have issues related to sun or water exposure.\\ \textbf{Answer}: Yes, genetic predisposition to dermographism may play a role in infants experiencing syncope during bathing, as it may be related to a pediatric form of aquagenic urticaria.
\end{mdframed}

\section*{Ethical Statement}
Our research on AuthentiGPT strictly adhered to ethical standards, with the objective of promoting safety and trust in the AI community. We recognize that AuthentiGPT does not solve all ethical challenges related to LLMs, advocating for ongoing interdisciplinary collaboration. Our commitment is to maintain ethical transparency and to constantly reassess our ethical considerations as the field of AI continues to evolve.

\section*{Acknowledgement}
We would like to thank OpenAI, GPTZero, and Originality.ai for providing their API services. We are grateful to Francis Huang for providing helpful feedback and discussion.

\end{document}